\documentclass[lettersize,journal,onecolumn]{IEEEtran}
\usepackage{amsmath,amsfonts}
\usepackage{algorithmic}
\usepackage{algorithm}
\usepackage{array}
\usepackage[caption=false]{subfig}
\usepackage{textcomp}
\usepackage{stfloats}
\usepackage{url}
\usepackage{verbatim}
\usepackage{graphicx}
\usepackage{cite}
\usepackage[utf8]{inputenc} 
\usepackage[T1]{fontenc}    
\usepackage{hyperref}       
\usepackage{url}            
\usepackage{booktabs}       
\usepackage{amssymb}
\usepackage{amsfonts}       
\usepackage{nicefrac}       
\usepackage{microtype}      
\usepackage{doi}
\usepackage{float}
\usepackage{xcolor}
\usepackage{amsmath, xparse}
\usepackage{orcidlink}
\usepackage{comment}

\title{Privacy-aware Berrut Approximated Coded Computing applied to general distributed learning} 
     
\author{ Xavier
   Mart\'{\i}nez-Lua\~na$^\text{\orcidlink{0000-0001-9066-983X}}$,
   Manuel Fern\'andez-Veiga$^\text{\orcidlink{0000-0002-5088-0881}}$~\IEEEmembership{Senior
     Member,~IEEE},
   Rebeca P. D\'{\i}az-Redondo$^\text{\orcidlink{0000-0002-2367-2219}}$,
   Ana Fern\'andez-Vilas$^\text{\orcidlink{0000-0003-1047-2143}}$
   \thanks{X. Mart\'{\i}nez-Lua\~na is with the Galician
     Research and Development Center in Advanced Telecommunications
(GRADIANT)
     Estrada do Vilar, 56-58, Vigo, 36214, Spain. Email:
xmartinez@gradiant.org}
   \thanks{X. Mart\'{\i}nez-Lua\~na, M. Fern\'andez-Veiga and R. P. D\'{\i}az-Redondo
     are with atlanTTic, Information \& Computing Lab,
     Telecommunication Engineering School, Universidade de Vigo Vigo, 36310,
     Spain. Emails: xamartinez@alumnado.uvigo.gal, rebeca@det.uvigo.es,
     mveiga@det.uvigo.es}
}
\begin{document}

\maketitle

\begin{abstract}
Coded computing is one of the techniques that can be used for privacy protection in Federated Learning. However, most of the constructions used for coded computing work only under the assumption that the computations involved are exact, generally restricted to special classes of functions, and require quantized inputs. This paper considers the use of Private Berrut Approximate Coded Computing (PBACC) as a general solution to add strong but non-perfect privacy to federated learning. We derive new adapted PBACC algorithms for centralized aggregation, secure distributed training with centralized data, and secure decentralized training with decentralized data, thus enlarging significantly the applications of the method and the existing privacy protection tools available for these paradigms. Particularly, PBACC can be used robustly to attain privacy guarantees in decentralized federated learning for a variety of models. Our numerical results show that the achievable quality of different learning models (convolutional neural networks, variational autoencoders, and Cox regression) is minimally altered by using these new computing schemes, and that the privacy leakage can be bounded strictly to less than a fraction of one bit per participant. Additionally, the computational cost of the encoding and decoding processes depends only of the degree of decentralization of the data.
\end{abstract}

\section{Introduction}
\label{sec:introduction}

Coded computing has recently received attention as an effective solution to solve 
privacy and security problems in centralized and decentralized computing systems, 
especially in machine learning frameworks~\cite{Kim2019,Li20,Ulukus22,BACC:NA20,Yu2020}. 
By using computations in the coded domain, these systems can work robustly against 
the presence of servers or clients that delay their responses or fail to finish a 
local computing task~\cite{Yu2020} (straggler nodes), can be protected against 
adversarial servers that corrupt the information that the learning algorithm 
uses~\cite{Pillutla2022}, and can enforce privacy guarantees to 
the input and output values in multi-party computation schemes~\cite{Yu19,Liu2023}.

Most of the existing coded computing techniques seek exact recovery of the outputs,
where one or more servers attempt to obtain the exact value of some function from 
its encoded arguments. Examples of broad classes of functions that are amenable to
this requirement of perfect recovery are typically structured computing tasks, like matrix 
multiplications~\cite{Dutta2020,Chen2020,Devulapalli2022,Kim2023,CodedSketch:21} or 
polynomial  evaluations~\cite{Raviv2019,Yu2017,Fahim2021}. However, these methods
operate with the restriction that the input/output values must belong to a certain
finite field, and additionally suffer from a crisp threshold for recovery, since
exact values can be calculated only if the number of stragglers or malicious nodes is 
below a limit. Above such limit, the computing task fails and no useful value
is produced~\cite{Yu2017}.
 
These limitations are not well matched to the requirements of current distributed
machine learning applications, in which many algorithms rely on computations lacking 
a specific structure or a decomposition property that simplifies the task. In these
contexts, additionally, the computations often involve non-linear functions (e.g.,
ReLU units in a neural network~\cite{Aly22}) not supported by classical coded 
computing methods, and operate over real or complex numbers directly instead of on 
discrete or quantized samples. Moreover, when computations are performed over the 
real numbers domain, approximate values for both the input and output values suffice, 
because the algorithms are iterative and not particularly sensitive to small errors in
each round~\cite{Liu2023}. In response to this, approximate coded computing has 
emerged  as a generalized approach to handle computing problems over a wider class of 
real- and complex-valued functions and, with the relaxation consisting of
performing only approximate computations, to reduce significantly the amount of
computing resources needed for completion of the task~\cite{Leon2025a,Leon2025}. 
While approximate recovery introduces errors in the computed function values, this 
error vanishes as the number of honest or non-straggler nodes increases, which is 
beneficial and practical for large distributed computing  applications composed of 
hundreds of nodes. 

Two approaches, Berrut Approximated Coded Computing (BACC~\cite{BACC:NA20}) and 
Learning Theoretic Coded Computing (LeTCC~\cite{Moradi2024}) have been proposed in 
the recent literature as general approximate coded computing solutions. BACC uses a 
barycentric approximation to evaluate the target function and has an approximation 
error that decreases quadratically in the number $n$ of honest nodes. LeTCC, in 
contrast, adopts an original approach in which the encoding and decoding functions 
used for computations and recovery are learned under the constraint of minimizing a 
loss function. As a consequence of being formulated as an optimization problem, 
its approximation error is lower,  it decreases as $n^{-3}$. 

However, neither BACC nor LeTCC  provide input or output privacy guarantees to the
approximately computed values. In our previous work~\cite{PBACCFL24} we extended BACC to
support privacy, where the privacy metric is  specified as a bound on the mutual
information between the encoded inputs and the output values. Yet, although our PBACC 
(Private BACC) provides input privacy in federated learning (FL) scenarios, this
is not sufficient for exploiting the advantages of generalized coded computing in
other important distributed computing systems, like in distributed learning with or 
in learning over decentralized private data. In this paper, we solve the integration 
of PBACC in this broader setting, focusing on decentralized federated learning
architectures with privacy-preserving training and aggregation. Specifically,
\begin{itemize}
\item We extend and adapt PBACC to distributed computing tasks where data is decentralized, 
i.e., there are many data owners who in addition wish to preserve privacy on their
data. The improved PBACC scheme can therefore be used for secure aggregation and for 
secure training in distributed learning systems.

\item PBACC is then integrated within Convolutional Neural Networks (CNN) and 
Variational Auto Encoders (VAE), including the loss function and gradient calculation. 
Thus,  we demonstrate the usefulness of the scheme for arbitrary function evaluations.

\item Our extensive numerical experiments show that PBACC can actually be leveraged in 
a variety of distributed computing cases, both for centralized and decentralized
data, and for secure aggregation of models or secure training at the clients as well. 
It also works efficiently for different machine learning models. The experimental
tests comprise a variety of network configurations, where the main threat comes from
a fraction of honest but curious nodes.
\end{itemize}

The rest of the paper is organized as follows. Section~\ref{sec:related_work} 
briefly reviews some relevant literature related to our work. Section~\ref{sec:cc} 
describes PBACC, the selected scheme to be integrated in the distributed
learning scenarios, and this is generalized in Section~\ref{sec:generalizing_pbacc}
to make it feasible and practical for different types of configurations and ML models. 
Section~\ref{sec:integrating_pbacc} presents in detail the different versions of
our scheme as they have to be used into centralized aggregation, decentralized 
aggregation, and decentralized training. This is followed by a discussion
on their privacy properties and  the form that PBACC should be integrated in 
each case, in Section~\ref{sec:comparison}. In Section~\ref{sec:results}, we describe 
all the tests performed for PBACC, and we discuss the results obtained. To finish, 
Section~\ref{sec:conclusions} summarizes the main conclusions of this work, and 
outlines some future work.

\section{Related work}
\label{sec:related_work}

Private Coded Distributed Computing (PCDC) refers to a subset of methods within the Coded Distributed Computing (CDC)
framework~\cite{Ng20}, developed to incorporate randomness to
ensure a certain level of privacy and security. These methods play a crucial role in mitigating key challenges in distributed learning~\cite{FederatedLearning:KGN23},
such as communication overhead, straggler issues, and privacy risks~\cite{Li20}. Lagrange Coded Computing (LCC)~\cite{Yu19} is a prominent method in the PCDC domain that offers a unified approach to evaluate general multivariate polynomial functions. LCC leverages the Lagrange interpolation polynomial to introduce redundancy in computations. It is resilient to stragglers, secure against malicious actors, and ensures input privacy while minimizing storage, communication, and randomness overheads. Nevertheless, LCC faces significant drawbacks: (i) it cannot handle ML activation functions, (ii) it exhibits numerical instability when working with rational numbers or in networks with a considerable number of nodes, and (iii) it requires input quantization into a finite field.

Analog LCC (ALCC)~\cite{Soleymani21} emerges as a solution to enable the application of LCC in the analog domain, but it does not resolve the other interpolation-related limitations. It is the Berrut Approximated Coded Computing (BACC)~\cite{BACC:NA20} which tackles the challenges of both LCC and ALCC in a different way. BACC enables the approximate computation of arbitrary target functions by distributing tasks across a potentially unlimited number of nodes, maintaining a bounded approximation error. 

Unfortunately, BACC lacks any privacy guarantee, this is why we developed PBACC~\cite{PBACC24}. PBACC extends BACC to include input privacy and generalizes the scheme for configurations with multiple data owners, making it suitable for distributed learning systems such as FL and decentralized FL. A critical feature of PBACC is its ability to compute
non-linear functions while balancing privacy, precision, and complexity.

There are alternative approaches to the already mentioned ones that were specifically
designed to the machine learning field. Some authors propose an optimal linear code for
private gradient computations~\cite{HarmonicCoding19} or a secure aggregation framework
that leverages Lagrange Coding~\cite{So21} that is able to break the quadratic barrier
of aggregation in Federated Learning. Within the FL field, there is an
approach~\cite{CFL21} that proposes applying Coded Federated Learning to mitigate the
impact of stragglers in Federated Learning, and another, coined as
CodedFedl~\cite{CodedFedL21}, that enables non-linear federated learning by efficiently
exploiting distributed kernel embeddings. Another work focused distributed learning
is~\cite{Soleymani22}, which proposes Analog Secret Sharing for the private distributed
training of a machine learning model in the analog domain. More
recently,~\cite{Moradi2024} explores optimal strategies for designing encoding and
decoding schemes in broader machine learning settings. Specifically, the problem of
designing optimal encoding and decoding functions is treated as a learning problem, with
the clear advantage that the resulting encoder/decoder pair is adapted to the statistical
distribution of the data. 

However, \cite{HarmonicCoding19} works only for gradient-type function and uses 
incremental redundancy that grows in harmonic progression, i.e., it cannot be used 
with other machine learning models and increases significantly the length of the
messages between the master and the workers. \cite{So21} can provide privacy for
aggregation at the master, not for general computations, and~\cite{CFL21,CodedFedL21}
are schemes resilient to stragglers but without privacy guarantees. Secret evaluation
of polynomials  over the real or complex numbers is the focus in~\cite{Soleymani22}, but
this work leaves out the problem of computing other types of functions.

\section{Background: Coded-Computing with Privacy-aware Berrut Approximation}
\label{sec:cc}

Our previous research work, Privacy-aware Berrut Approximated Coded Computing
(PBACC)~\cite{PBACC24}, has shown promising results in FL settings using a Convolutional
Neural Network (CNN) and two aggregation strategies (FedAvg and
FedMedian)~\cite{PBACCFL24}. In this paper, we go a step further to provide a suitable
solution for any decentralized machine learning scenario, which we have coined as
Generalized PBACC.

In order to include privacy in the BACC scheme, we work with the following threat model.
It is assumed that from the $N$ total worker nodes, up to $c$ nodes can be honest 
but curious. Curious nodes respect the computation protocol, so they are practically
impossible to detect, but they can exchange messages to collaboratively work to disclose
information (colluding behaviour). This implies that these nodes attempt to infer
information on the private data, denoted by $\mathbf{X}$, by observing the encoded
information received $\mathbf{Y}$. Under this scenario, the objective is PBACC be
$\epsilon$-secure under a privacy metric denoted as $i_L$, i.e. that PBACC be
$i_L \leq \epsilon$. 

In this background section, we firstly overview the original BACC scheme
(Sect.~\ref{sec:bacc}), in which we based our previous research work to add privacy:
PBACC, summarized in Sect.~\ref{sec:pbacc}. Finally, we describe in
Sect.~\ref{sec:measuring_privacy} the privacy metric $i_L$ we have defined to assess
the proposal.

\subsection{BACC: Berrut Approximated Coded Computing}
\label{sec:bacc}

BACC works on a network of one master node (owner of the data) and $N$ worker nodes 
that are in charge of computing an objective function $f: \mathbb{V} \rightarrow  
\mathbb{U}$ over some input data $\mathbf{X} = (X_0, \dots, X_{K-1})$, where 
$\mathbb{U}$ and $\mathbb{V}$ are vector spaces of real matrices. BACC approximately
computes $\tilde{f}(\mathbf{X}) \approx \bigl(f(X_1), \dots, f(X_{K - 1})\bigr)$, with 
a bounded error. This scheme is numerically stable, as it provides a result even in
scenarios with high number of nodes $N$ and $K$. It also resists stragglers, as the 
error depends on the number of received results, and it allows to approximate any
arbitrary function $f$ under some conditions (this will  be further explained in 
Sect.~\ref{sec:integrating_pbacc}). The BACC protocol works in three stages or phases, 
as follows: \\

\noindent {\textbf{[1] Encoding and Sharing}}. To perform the encoding operation over the 
input data $\mathbf{X}$, the master node computes the rational function 
$u: \mathbb{C} \rightarrow \mathbb{V}$, defined as
\begin{equation}
\label{eq:bacc}
    u(z) = \sum_{i = 0}^{K - 1} \frac{\frac{(-1)^i}{(z - \alpha_i)}}{\sum_{j = 0}^{K - 1} 
    \frac{(-1)^j}{(z - \alpha_j)}} X_i, 
\end{equation}
for some distinct interpolation points $\boldsymbol\alpha = (\alpha_0, \dots, 
\alpha_{K-1}) \in \mathbb{R}^K$. It is straightforward to verify that this mapping
is exact, $u(\alpha_j) = X_j$, for $j \in \{0, \dots, K - 1\}$. As per~\cite{BACC:NA20}, 
these decoder mapping points $\boldsymbol\alpha$ are selected as the Chebyshev 
points of first kind
\begin{equation}
\label{eq:chebychev-1st}
\begin{aligned}
  &\alpha_j = \operatorname{Re}\left\{\cos\left(\frac{(2j + 1) \pi}{2K}\right) + \imath 
  \sin\left(\frac{(2j + 1) \pi}{2K}\right)\right\},\\
  &j = 0, \dots, K-1
\end{aligned}
\end{equation}
where $\imath = \sqrt{-1}$. Then, the master node selects another set of $N$ 
distinct encoder mapping points $\boldsymbol\beta = \{\beta_0, \dots, \beta_{N-1} \}$,
computes each share $u(\beta_j)$, and sends it to worker $j$. In~\cite{BACC:NA20}, 
it is suggested to choose $\{ z_j: j = 0, \dots, N-1 \}$ as the Chebyshev points of 
second kind
\begin{equation}
\label{eq:chebychev-12nd}
\begin{aligned}
  &\beta_j = \operatorname{Re}\left\{\cos\left(\frac{j\pi}{N-1}\right) + \imath 
  \sin\left(\frac{j\pi}{N-1}\right)\right\}, \\
  &j = 0, \dots, N-1.
\end{aligned}
\end{equation}

\noindent {\textbf{[2] Computation and Communication}}. Each worker $j$ receives the 
share $u(\beta_j)$, and computes the result $v_j = f\bigl( u(\beta_j) \bigr)$ applying 
the target function $f(\cdot)$, for $j \in \{0, 1, \dots, N-1\}$. Then, each worker 
$j$ sends the result $v_j$ to the master node. \\
    
\noindent {\textbf{[3] Decoding}}. When the master collects $n \leq N$ results from 
the  subset $\mathcal{F}$ of fastest nodes, it approximately calculates $f(X_j)$, for 
$j = \{0, \dots, K -  1 \}$, using the decoding function based on the Berrut 
rational  interpolation
\begin{equation}
\label{eq:berrut-decoding}
  r_{\mathrm{Berrut}, \mathcal{F}}(z) = \sum_{i = 0}^{n} \frac{\frac{(-1)^i}{(z - 
  \tilde{\beta}_i)}} {\sum_{j = 0}^{n}\frac{(-1)^j}{(z - \tilde{\beta}_j)}} f(u(\tilde{\beta}_i)),
\end{equation} 
where $\tilde{\beta}_i \in \mathcal{S}$ are the evaluation points 
$\mathcal{S} = \{ \cos\frac{j\pi}{N-1}, j \in \mathcal{F} \}$ corresponding to the 
$n$ faster nodes. The result of this decoding operation is the approximation 
$f(X_i) \approx r_{\mathrm{Berrut}, \mathcal{F}}(\alpha_i)$, for $i \in \{0, \dots, 
K - 1 \}$.

\subsection{PBACC: Privacy-aware Berrut Approximated Coded Computing}
\label{sec:pbacc}

Our objective to add privacy in the BACC scheme found a critical challenge: since 
it deals with rational functions, it cannot achieve perfect information-theoretical
privacy, where the adversaries  cannot learn anything about the local input. 
Therefore, our objective was change to achieve a bounded information leakage lower 
than $\epsilon$, a target security parameter. This value represents the maximum amount 
of leaked information per data point that is allowed for a fixed number of colluding 
semi-honest nodes $c$. The PBACC protocol we proposed~\cite{PBACC24} works in three 
stages or phases, exactly as the BACC protocol. In fact, stage [2] Computation and
Communication and stage [3] Decoding, are the same as it was previously detailed 
(Sect.~\ref{sec:bacc}). The changes are only located in the first step, as follows: \\

\noindent {\textbf{[1] Encoding and Sharing}} To perform the encoding of the input 
data $\mathbf{X}$, the following rational function $u: \mathbb{C} \rightarrow 
\mathbb{V}$ is composed by the master node
\begin{equation}
\label{eq:pbacc}
  \begin{aligned}
        u(z) &= \sum_{i =0}^{K - 1} \frac{\frac{(-1)^{i}}{(z - \alpha_i)}}{\sum_{j = 0}^{K + T -1} 
        \frac{(-1)^j}{(z - \alpha_j)}} X_i + \sum_{i = 0}^{T - 1} \frac{\frac{(-1)^{K+i}}{(z - \alpha_{K + 
        i})}}{\sum_{j = 0}^{K + T - 1}\frac{(-1)^j}{(z - \alpha_j)}} R_i\\
        &= \sum_{i = 0}^{K + T - 1} \frac{\frac{(-1)^{i}}{(z - \alpha_i)}}{\sum_{j = 0}^{K + T - 1}\frac{(-1)^j}{(z - 
        \alpha_j)}} W_i,
    \end{aligned}
\end{equation}
where $\mathbf{W} = (X_0, \dots, X_{K-1}, R_0, \dots, R_{T-1}) = 
(W_0, \dots, W_{K + T - 1})$. Here, $\{R_{i}: i = 1, \dots, T - 1 \}$ are random 
data points independently generated according to a Gaussian distribution
$\mathcal{N}(0, \frac{\sigma_n^{2}}{T})$ with zero mean and variance 
$\frac{\sigma_n^{2}}{T}$, for some distinct points $\boldsymbol\alpha = (\alpha_0, 
\dots, \alpha_{K+T-1}) \in \mathbb{R}^{K+T}$. The data decoder mapping points of $X$ 
are  chosen again as Chebyshev points of first kind
\begin{equation}
  \alpha_j = \operatorname{Re}\left\{\cos\left(\frac{(2j + 1) \pi}{2K}\right) + \imath 
  \sin\left(\frac{(2j + 1) \pi}{2K}\right)\right\},
\end{equation}
for $j \in \{0, \dots, K-1\}$, whereas the decoder mapping points of $R$ are 
chosen as shifted Chebyshev points of the first kind
\begin{equation}
\label{eq:shifted-chebyshev-1st}
  \alpha_j = b + \operatorname{Re}\left\{\cos\left(\frac{(2j + 1) \pi}{2T}\right) + \imath 
  \sin\left(\frac{(2j + 1) \pi}{2T}\right)\right\},
\end{equation}
where $b \in \mathbb{R}$, for $j = 0, \dots, T - 1$. By definition, it also holds
that decoding is exact $u(\alpha_j) = X_j$, for $j \in \{0, \dots, K - 1\}$. Then, 
the master node selects $N$ distinct points $\boldsymbol\beta = \{ \beta_0, \ldots,
\beta_{N - 1} \}$, computes $u(\beta_j)$, and assigns this value to worker $j$. 
The encoding mapping points $\boldsymbol\beta$ are chosen as the Chebyshev points 
of second kind
\begin{equation}
  \beta_j = \operatorname{Re}\left\{\cos\left(\frac{j\pi}{N-1}\right) + \imath \sin\left(\frac{j\pi}{N-1}\right)\right\}, \quad 0 \leq j \leq N-1.
\end{equation}

\subsection{Privacy metric}
\label{sec:measuring_privacy}

We defined a privacy metric~\cite{PBACC24} based on the worst-case achievable
mutual information for the subset of colluding nodes $c$. In order to bound this 
mutual information $I_L$, we leverage on results about the capacity of a 
Multi-Input Multi-Output (MIMO) channel under some specific power 
constraints~\cite{MIMOCapacity:SPM02}. From an information-theoretic perspective, 
a MIMO channel with $K$ transmitter antennas and $c$ receiver antennas is equivalent 
to a signal composed of $K$ encoded element of the input, and a cooperative observation 
of this signal by the $c$ semi-honest colluding nodes. Given this conceptualization, the 
encoded noise of the scheme provides privacy, as it reduces the capacity of the 
channel, which implies that less information is received by the $c$ colluding nodes.

Assuming that the noise introduced by PBACC is Gaussian, we have therefore an 
Additive White Gaussian Noise (AWGN) vector channel. This allows to bound the 
mutual information using known results on the capacity of a MIMO channel with 
correlated noise and uniform power allocation~\cite{MIMOCapacity:SPM02},
\begin{equation}
\label{eq:C}
    C = \sup_{P_{\mathbf{X}}} I(\mathbf{Y}; \mathbf{X}) = \log_2 | I_{c} + P H 
    \Sigma^{-1}_{\mathbf{Z}} H^{\dagger} |,
\end{equation}
where $\mathbf{X}$ is the private input, $\mathbf{Y}$ is the encoded output, $P$ 
is the maximum power of each transmitter antenna, $I_c$ the identity matrix of 
order $c$ and $|\cdot|$ the determinant of a matrix.

By the definition of $I_L$ (worst-case achievable for $c$ colluding nodes), we can 
define it as
\begin{equation}
    I_L \triangleq \max_{\mathcal{C}} \sup_{P_\mathbf{X}: ||X_i| \leq s|, \forall i \in [K]}  
    I(\mathbf{Y}; \mathbf{X}),
\end{equation}
where $P_\mathbf{X}$ is the probability density function of $\mathbf{X}$, $s$ 
determines the interval $D_s \triangleq [-s, s]$ from which the input random variable 
can take values, and the maximization applies to any set of colluding nodes 
$\mathcal{C} \subset [N]$. As $\|X_i\| \leq s$, the power $\mathbb{E}[ \|X_i\| ]^2 
\leq s^2$. Thus, the previous equation can be re-written as
\begin{equation}
  \label{eq:I_L_1}
    I_L \leq \max_{\mathcal{C}} \sup_{P_\mathbf{X}: \mathbb{E}[ \|X_i \|^2] \leq s^2}  
    I(\mathbf{Y}; \mathbf{X}).
\end{equation}
Combining~\eqref{eq:C} and~\eqref{eq:I_L_1}, and assuming the noise is uncorrelated, 
we can write
\begin{equation}
  \label{eq:updated_eta_c}
    I_L \leq \max_{\mathcal{C}} \log_2 \bigl| I_C + \frac{s^2 T}{\sigma^2_n} \tilde{\Sigma}_C^{-1} \Sigma_C \bigr|,
\end{equation}
where $I_C$ is the identity matrix of size $C \times C$, $T$ is the number of 
random coefficients, $\sigma_n$ the standard deviation of the noise, and 
$\tilde{\Sigma}_C$ and $\Sigma_C$ are the covariance
matrices of the encoded input and noise respectively. These matrices are defined as
\begin{equation}
\begin{aligned}
 \Sigma_{c} &\triangleq
    \begin{pmatrix}
        q_{0}(\beta_{i_1}) & \hdots & q_{K-1}(\beta_{i_1}) \\
        q_{0}(\beta_{i_2}) & \hdots & q_{K-1}(z_{\beta_2}) \\
        \vdots & \ddots & \vdots \\
        q_{0}(\beta{i_{c}}) & \hdots & q_{K-1}(\beta_{i_{c}})
    \end{pmatrix}_{c \times K}, \\
  \tilde{\Sigma}_{c} &\triangleq 
    \begin{pmatrix}
        q_{K}(\beta_{i_1}) & \hdots & q_{K+T-1}(\beta_{i_1}) \\
        q_{K}(\beta_{i_2}) & \hdots & q_{K+T-1}(\beta_{i_2}) \\
        \vdots & \ddots & \vdots \\
        q_{K}(\beta_{i_{c}}) & \hdots & q_{K+T-1}(\beta_{i_{c}})
    \end{pmatrix}_{c \times T},
\end{aligned}
\end{equation}
where
\begin{equation}
    q_i(z) = \frac{\frac{(-1)^i}{(z-\alpha_i)}}{\sum_{j=0}^{K+T-1}\frac{(-1)^j}{(z-\alpha_j)}},
\end{equation}
and $\{ \beta_{i_h} \}$ are the evaluation points of the $c$ colluding nodes.

The final privacy metric is defined as the normalized $i_L = \frac{I_L}{K}$, that 
denotes the maximum information leakage per data element in presence of $c$ 
colluding nodes. Therefore, PBACC is $\epsilon$-secure if $i_L \leq \epsilon$.

\section{Generalized PBACC}
\label{sec:generalizing_pbacc}

Since our purpose is generalizing the PBACC scheme for any computation configuration, we need to tackle two aspects: (i) the number of input sources, and (ii) the input data. The former entails the system would operate for multiple data owners and was already solved in our previous research work~\cite{PBACC24}. The latter entails accepting tensors instead of matrices, which is essential to support encoding any parameter or dataset, is the one we describe in this section.

We assume a set of $N$ nodes with their own private data tensor
\begin{equation}
  \mathbf{X}^{(j)}_{k_0 k_1 \ldots k_{L-1}} =
  \begin{pmatrix}
    X_{0 k_1 \ldots k_{L-1}}^{(i)}\\
    \vdots\\
    X_{K-1 k_1 \ldots k_{L-1}}^{(i)}\\
  \end{pmatrix}
\end{equation}
where the input data $\mathbf{X}^{(i)} \in \mathbb{R}^{K\times k_1 \ldots 
\times k_{L-1}}$ is owned by node $i$, for $i \in \{0, \dots, N-1\}$, and $L$ denotes 
the rank of the tensor. Note that we fix $k_0 = K$, to indicate that the 
scheme operates (and compresses) the first dimension of the tensor, but any other
dimension can used as well. The generalized scheme consists of three phases, that 
are detailed  in the following. \\

\noindent\textbf{[1] Encoding and Sharing}. The client node $i$ composes the following 
encoded polynomial quotient
\begin{equation}
  \label{eq:generalized_pbacc}
  \begin{aligned}
    u_{\mathbf{X}^{(i)}}(z) = &\sum_{j = 0}^{K - 1} \frac{\frac{(-1)^j}{(z - \alpha_j)}}{\sum_{k = 0}^{K + T - 1} \frac{(-1)^k}{(z - \alpha_k)}} X_{j k_1 \ldots k_{L-1}}^{(i)}
        +\\  &\sum_{j = 0}^{T-1} \frac{\frac{(-1)^{j + K}}{(z -\alpha_{j})}}{\sum_{k = 0}^{K + T - 1}\frac{(-1)^j}{(z - \alpha_k)}} R_{j k_1 \ldots k_{L-1}}^{(i)},
   \end{aligned}
\end{equation}
for some distinct interpolation points associated to the the data 
$\mathbf{X}_{k_0 k_1 \ldots k_{L-1}}^{(i)}$, data decoder mapping points 
$\boldsymbol\alpha = (\alpha_0, \dots, \alpha_{K-1}) \in \mathbb{R}^K$, which we 
choose  also in this case as the Chebyshev points of first kind  
$\alpha_j = \cos\left(\frac{(2j + 1)\pi}{2K}\right)$, and
distinct noise encoder mapping points associated to the randomness 
$\mathbf{R}_{T k_1  \ldots k_{L-1}}^{(i)}$, $\alpha_K, \dots, \alpha_{K + T - 1}  
\in \mathbb{R}$, which are selected as the shifted Chebyshev points  of first kind 
$\alpha_{K + j} = b + \cos\left(\frac{(2j + 1)\pi}{2T}\right)$. The 
rational function $u_{\mathbf{X}^{(i)}}(z)$ is evaluated at the set
of encoder mapping points $\boldsymbol\beta = \{\beta_j \}$, for 
$j = 0, \dots, N-1$, using $\beta_j = \cos(\frac{j\pi}{N - 1})$, the Chebyshev 
points of second kind. Finally, the random coefficients $R_{j k_1 \ldots k_{L-1}}^{(i)}$
are  distinct tensors drawn from a Gaussian distribution 
$\mathcal{N}(0,  \frac{\sigma_n^2}{T})$. The evaluation $u_{\mathbf{X}^{(i)}}(\beta_j)$ 
is the share created from the client node $i$ and sent to the node $j$, as explained 
below. \\

\noindent\textbf{[2] Computation and communication}. In this phase, node $i$ 
calculates an arbitrary function $f$ using a set of  polynomial quotient 
evaluations shared from the rest of nodes $\{ u_{\mathbf{X}^{(0)}}(\beta_i)$, 
$u_{\mathbf{X}^{(1)}}(\beta_i), \dots, u_{\mathbf{X}^{(N-1)}}(\beta_i)\}$, where
$u_{\mathbf{X}^{(j)}}(\beta_i)$ is the evaluation of the rational function 
corresponding to the input $X^{(j)}$ owned by the $j$-th node, and shared with 
node $i$.  The client node $i$ computes $f\bigl( u_{\mathbf{X}^{(0)}}(\beta_i), 
u_{\mathbf{X}^{(1)}}(\beta_i), \dots, u_{\mathbf{X}^{(N-1)}}(\beta_i) \bigr)$ 
and sends the result to the master node. \\

\noindent\textbf{[3] Decoding}. In this last phase, the master node reconstructs 
the value of desired function over all the inputs using the results obtained from 
the subset $\mathcal{F}$ of the fastest workers, computing the reconstruction 
function
\begin{multline}
\label{eq:reconstruction}
    r_{\mathrm{Berrut}, \mathcal{F}}(z) =
    \sum_{i = 0}^{n} \frac{\frac{(-1)^i}{(z - 
    \tilde{\beta}_i)}}{\sum_{j = 0}^{n} \frac{(-1)^j}{(z - \tilde{\beta}_j)}} f\bigl( u_{\mathbf{X}^{(0)}}(\tilde{\beta}_i),
    u_{\mathbf{X}^{(1)}}(\tilde{\beta}_i),\dots,\\ u_{\mathbf{X}^{(N-1)}}(\tilde{\beta}_i) \bigr),
\end{multline}
where $\tilde{\beta}_i \in \mathcal{S} = \{ \cos\frac{j\pi}{N-1}, j \in \mathcal{F} \}$
are the evaluation points of the fastest nodes. After this, the master node finds 
the approximation $f\bigl(X^{(0)}_{j k_1 \ldots k_{L-1}}$, $X^{(1)}_{j k_1 \ldots 
k_{L-1}}, \dots, X^{(N-1)}_{j k_1 \ldots k_{L-1}}\bigr) \approx r_{\mathrm{Berrut}, 
\mathcal{F}}(\alpha_j)$, for all $j \in \{0, \dots, K - 1\}$.


\section{Integrating PBACC in Decentralized Learning settings}
\label{sec:integrating_pbacc}


One of the criteria to classify decentralized learning scenarios is the owner of the data. When there is only one data owner (named as master), it distributes the computation, training the machine learning model, among a set of other nodes. Otherwise, when there are different datasets owned by different nodes, there is also a master role, adopted by one of the nodes, whereas the others are in charge of computing the machine learning training. In this case, the master can be leveraged for some specific tasks like obtaining
the global aggregated model. According to this classification, there are three different approaches to secure the decentralized learning settings:

\begin{enumerate}
\item \textbf{Secure the training phase when there is only one owner (master).} The data owner does not have enough computing resources or it want to speed up the computation. Thus, the training is delegated to a number of workers who must not learn any significant information about the data.
\item \textbf{Secure the  model or parameters aggregation task when there are multiple data owners.} Here, the training phase is locally performed by each owner using its own data, but the synthesis of the global model takes place at a central node. Neither honest-but-curious nodes nor malicious nodes must learn statistically significant information on this global model if they have access to some of the
local models.
\item \textbf{Secure the whole training phase when there is only one owner (master).} The master node reveals the global model under an encoded form, and the other nodes do the local training directly on the coded domain, i.e., without 
disclosing such global model, which is therefore kept private.
\end{enumerate}

We can categorize these three approaches and their corresponding unsecure (uncoded)
versions into two categories:
\begin{enumerate}
    \item Distributed Learning over Centralized Data (DLCD). This category comprises
    all the approaches where the master is the unique owner of the data, and shares it
    to the nodes to distribute the learning task.
    \item Distributed Learning over Dentralized Data (DLDD). This category comprises
    all the approaches where the nodes are the owners of the data, and the master
    shares the global model to distribute the learning task. Please, note that this is
    in fact Federated Learning, but we rename it to maintain a consistency with the
    other option.
\end{enumerate}

For simplicity, we use the following notation: the aggregation function is $\mathsf{agg}()$,
ML model parameters are denoted by $\boldsymbol\theta$, and the training function by
$F_{\boldsymbol\theta}()$. The latter can in turn be composed of one or many of the 
following functions for each of its steps: model computation, evaluation of the loss
function, calculation of gradients and model optimization. The degree of decentralization
in the training phase is higher if $F_{\boldsymbol\theta}$ subsumes a higher number of these
steps. So, we consider the possibility of computing multiple epochs of the local model before
sharing the result, which this would heavily reduce the communication costs. This introduces
a trade-off with precision, since computation of a more complex function induces lower model
accuracy, generally. The only assumption common to these three cases is that the model
training follows a typical mini-batch Gradient Descent configuration.

\subsection{DLCD: Distributed Learning over Centralized Data}

\begin{figure}[t]
\centering
\includegraphics[width=0.7\linewidth]{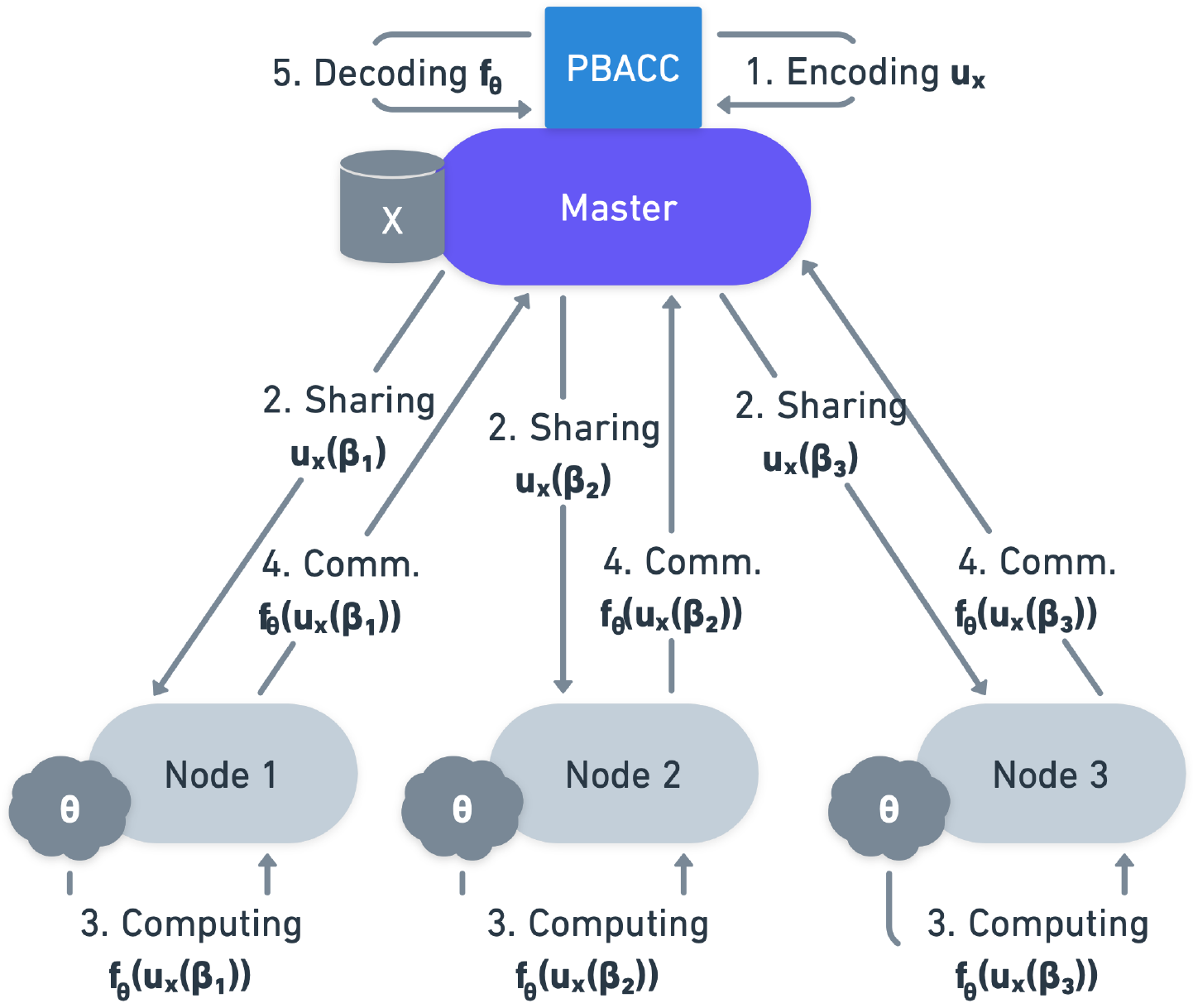}
\caption{Distributed training over centralized data}\label{centralized_data_secure_training}
\end{figure}

In this scenario there is only one data owner (master) of the whole 
dataset $X_{k_0 k_1 \ldots k_{L-1}}$ that wants to distribute the training of a 
machine learning model to a network of $N$ nodes.  The rank of 
the tensor $L$ will depend on the nature of the dataset (e.g., records, images, etc.) 
The master node splits the dataset into $N$ parts, and sends each $X_{k_0/N k_1 
\ldots k_{L-1}}^{(j)}$ along with the parameterized global model $f_{\boldsymbol\theta}$
to node $j \in \{0, 1, \dots, N-1\}$ ($\boldsymbol\theta$ denotes the vector of 
parameters for the function class). Each node $j$ computes its part of the training
obtaining a new model $f_{\boldsymbol\theta}^{(j)}$, and sends it back to the master. 
This entity aggregates all of the received model parameters into a new global model
$f_{\boldsymbol\theta}$. This process continues until convergence is achieved. 
The detailed steps on how it  has been secured with PBACC are explained next, and
are summarized in Figure~\ref{centralized_data_secure_training}. To achieve input
privacy, it suffices to encode the input dataset $X_{k_0 k_1 \ldots k_{L-1}}$ with
PBACC and let the nodes operate with their encoded versions of it.

Depending on wether we want to prioritize reducing
computation cost or communication cost, the encoding will be done 
differently. One option is to encode the whole dataset in a single 
interpolation point using~\eqref{eq:generalized_pbacc}, so $K=1$ and 
$\boldsymbol{\alpha} = \alpha_0$. This way of encoding will allow the worker
nodes to compute  the whole training task (model computation, loss function, 
gradient calculation and model update) by themselves, without any interaction 
with the master. Then, the master will receive the results from the nodes and 
reconstruct the new global model using~\eqref{eq:reconstruction}. This implies 
that the communication cost is very similar to the non-secured approach, but the 
nodes will have to compute the whole dataset in each round, so the computation cost
increases substantially. 

Another possibility is to split the 
dataset in batches of size $K$, and then encoding each element of the 
batch into a distinct  $\alpha_j = \{\alpha_0, \dots, \alpha_{K-1}\}$, 
using~\eqref{eq:generalized_pbacc}. Each node $j$ will receive an encoded version 
of the dataset $u_{X}(\beta_j)$ of reduced size $\frac{D}{K} \times 1 $, 
for $j \in \{0, \dots, N-1\}$. In this case, each node $j$ will just compute the 
model execution part of the training, and send the obtained result back to the master. 
The reason the nodes cannot continue with the rest of the execution is due to the 
loss function merging the results of the model execution into one single value. 
This implies that the encoding coefficients are merged together as well, so the 
result is no longer encoded into a known interpolation point, which makes
decoding impossible. Once the master has received all of the results corresponding 
to a batch, it will reconstruct the $K$ outputs of the model 
using~\eqref{eq:reconstruction}. Then, it will evaluate the 
loss function over the reconstructed results, calculate the gradients, update the
model, and send this updated version back to the nodes, thus they can continue
training with the next encoded batch. Obviously, this interaction between nodes and 
master after each model computation increases the communication costs. The greater 
the batch size is, less communication and computation cost is required, so 
this configuration will benefit from large networks and datasets, where large 
batch sizes can be chosen without affecting too much into the precision of the 
scheme. Compared to the other possibility, this one will heavily reduce computation
costs in exchange for communications costs. Hence, this will be the option evaluated
in Section~\ref{sec:results}.

\subsection{DLDD: Distributed Learning over Decentralized Data}

In a scenario with multiple owners (decentralized data) the input data is the one that needs to be protected. To this end, two distinct approaches can be followed: (i) 
secure the aggregation phase, or (ii) secure the training stage. Clearly, both options have their advantages and disadvantages, and choosing 
one or another depends on the specific network and on the application domain of the data.

Thus, there is a master node that owns the global model 
$f_{\boldsymbol\theta,k_0 k_1 \ldots k_{L-1}}$ and wants to distribute the training to a set of nodes. Each node $j$ receives the global model, and 
trains an updated one with its own private data batch 
$X_{k_0 k_1 \ldots k_{L-1}}^{(j)}$, for $j \in \{0, 1, \dots, N-1\}$.
Then, each node $j$ sends its updated model $f_{\boldsymbol\theta^{(j)},k_0 k_1 
\ldots k_{L-1}}$ back to the master, so this entity can aggregates all of the 
locally trained model parameters into the new global model. This process 
continues until convergence. The detailed steps on how PBACC can enforce
privacy in these two configurations are explained next.

\subsubsection{Secure aggregation over decentralized data}
\begin{figure*}[t]
    \centering
    \includegraphics[width=0.8\linewidth]{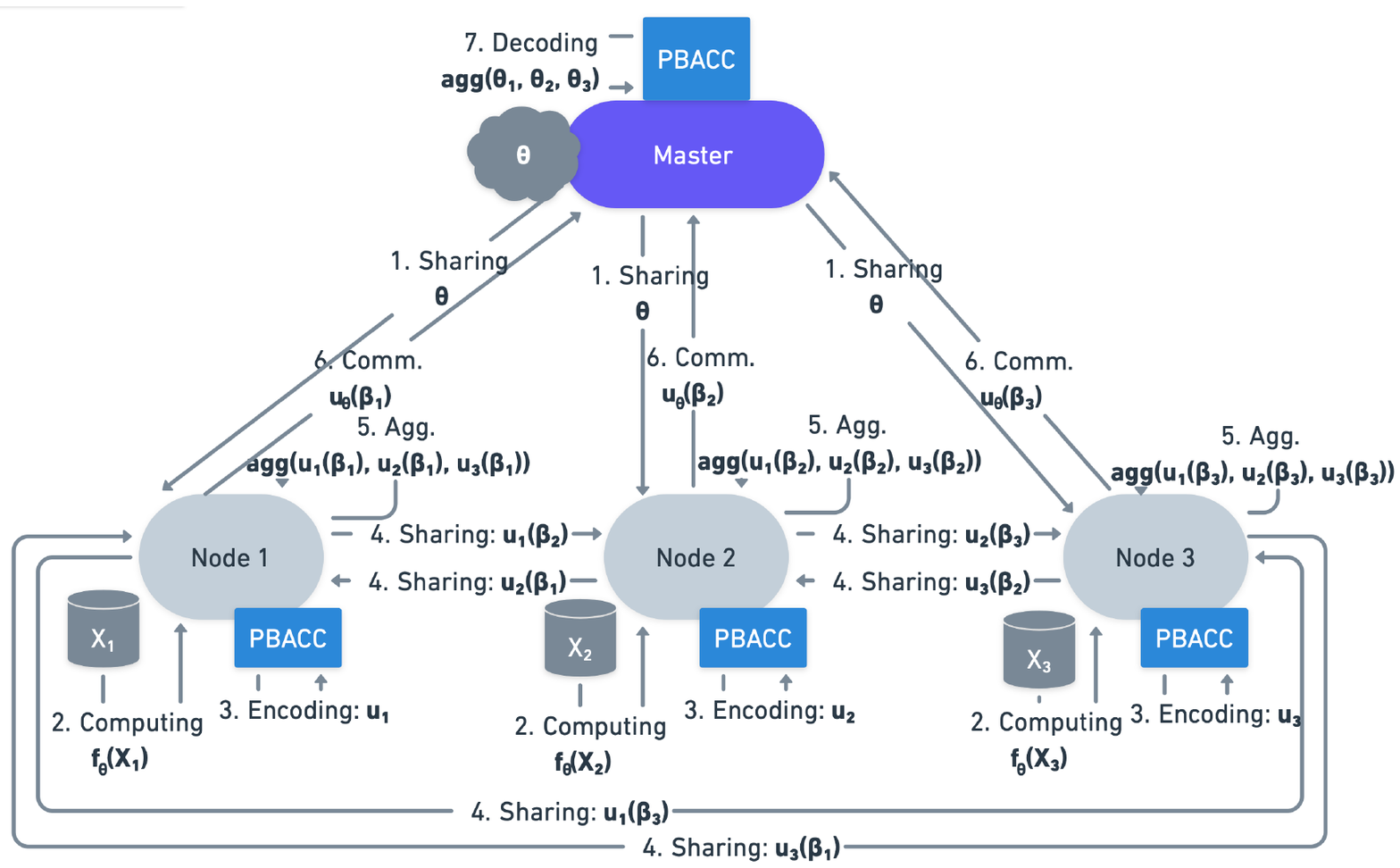}
    \caption{Secure aggregation over decentralized data}
    \label{fig:secure-aggregation}
\end{figure*}

In this case, the master sends the global model 
$f_{\boldsymbol\theta,k_0 k_1 \ldots k_{L-1}}$ to each node $j$, which then performs 
the training using its input dataset $X^{(j)}$, for $j \in \{0, 1, \dots, N-1\}$. 
This training task ends with each node $j$ obtaining its own version of the model 
$f_{\boldsymbol\theta^{(j)},k_0 k_1 \ldots k_{L-1}^{(j)}}$. To perform the aggregation 
in a secure way, each node $j$ encodes its model parameters into distinct 
$\alpha_j = \{\alpha_0, \dots, \alpha_{K-1}\}$, using~\eqref{eq:generalized_pbacc}, 
which results in $N$  shares $\mathbf{u}_j = [u_j(\beta_0), u_j(\beta_1), \dots,$ 
$u_j(\beta_{N-1})]$, that each of them is sent to a different node.
Each node $j$ ends up with $N$ shares $u_i(\beta_j)$, for $i, j \in \{0, 1, 
\dots, N-1\}$, aggregates them all using some known strategy $\mathsf{agg}(\cdot)$ 
(e.g.,  FedAvg~\cite{McMahan2016}, FedProx~\cite{Li2018},
SCAFFOLD~\cite{Karimireddy2020}, FedGen~\cite{Zhu2021}, etc.), and sends the 
result back to the master. The master collects the received results 
$\mathsf{agg}(u_1(\beta_j), \dots, u_{N-1}(\beta_j))$, for 
$j \in \{0, 1, \dots, N-1\}$, to reconstruct the new global model $f_{\boldsymbol\theta}$
using the decoding function~\eqref{eq:reconstruction}. The complete set of
operations is depicted in Figure~\ref{fig:secure-aggregation}.

This procedure to secure the aggregation stage increases the communication cost, 
as $N(N-1)$ additional messages will have to be sent in order to compute the 
aggregation. However, the shares have a reduced size 
$\frac{k_0}{K} k_1 \ldots k_{L-1}$, where $K$ the number of distinct decoder mapping 
points $\boldsymbol{\alpha}$ used to encode the data. Increasing $K$ would 
decrease the communication cost in exchange for precision on the aggregation, 
since more decoder mapping points will have to be processed. This factor $K$ would
also affect the aggregation cost, which decreases in $K$.

\subsubsection{Secure Training over decentralized data}

\begin{figure}[t]
    \centering
    \includegraphics[width=0.7\linewidth]{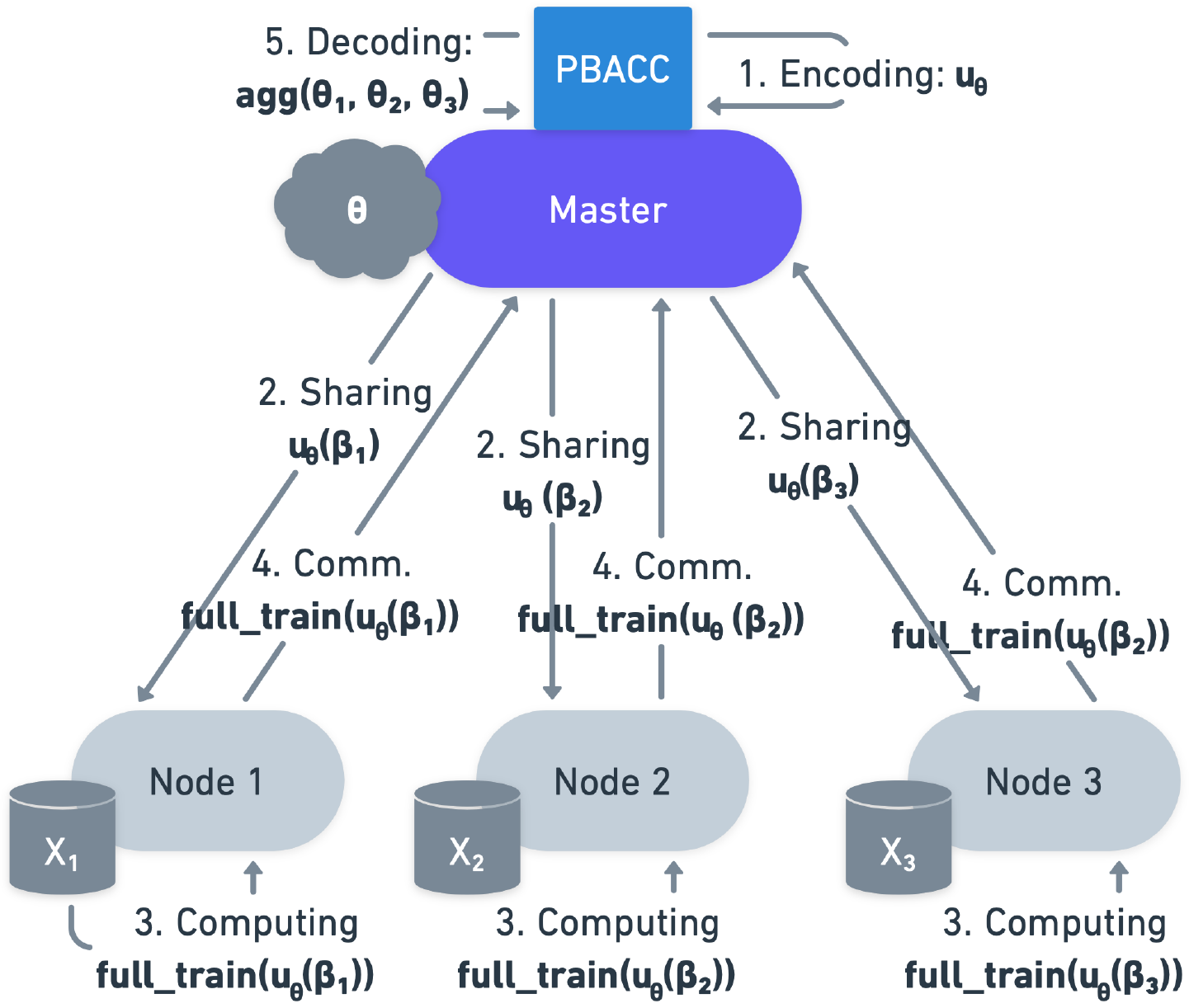}
    \caption{Secure Training over decentralized data}
    \label{fig:secure-training}
\end{figure}

In this case, the master encodes the global model 
$f_{\boldsymbol\theta,k_0 k_1 \ldots k_{L-1}}$ into one single interpolation point 
$\alpha$, using~\eqref{eq:generalized_pbacc}, resulting in $N$ shares
$\mathbf{u}_{\theta} = [u_{\theta}(\beta_0), u_{\theta}(\beta_1),$ $\dots,u_{\theta}(\beta_{N-1})]$,
and the master sends each of them to a different node. Then, each worker $j$ performs the 
whole training task using its encoded model $u_{\theta}(\beta_j)$, and its own private 
dataset $X^{(j)}$. This training function includes the model computation, the
loss function, the gradient calculation, and the model update. Finally, each node 
$j$ sends the encoded result $\mathsf{full\_train}(u_{\theta}(\beta_j))$ back to the master. 
This entity collects all results and decodes the new global model 
$f_{\boldsymbol{\theta^\prime}}$ using the reconstruction 
function~\eqref{eq:reconstruction}. As this function is not the same for every worker
$j$, this would appear to be a problem but actually, since the Berrut rational 
interpolation~\eqref{eq:reconstruction} is $\delta$-stable, and is based on the
barycentric interpolation, the resulting decoded output is theoretically
similar  to an average of all the trained models.

The reason behind the model being encoded into one single interpolation point relies 
on the fact that we are not dealing with model splitting in this work. Since each 
node  trains a local model of the same size as the global one, if the encoded version 
uses more than one interpolation point, these will be merged together during the 
calculation of the loss function, and will propagated back to the model parameters 
during the model  update step. Again, this would hinder the decoding step, 
because the data to be interpolated is not in the expected decoder mapping points 
anymore. Since all the models have the same size, even if it was possible to encode 
the model at various points, we could not compress the shares sent, for that would 
alter the  model size. The improvement would be in the sharing phase, where the
results would get compressed by a factor $K$.

\section{Comparison of PBACC decentralized learning models}
\label{sec:comparison}

Once the different operation models have been described, we provide a comparison according to privacy and efficiency from a theoretical perspective. The privacy analysis considers input privacy,
taking into account who owns the data and if it reveals it to another entity. The efficiency analysis has been focused on each round of the protocol, putting together the computation and the communication costs.

\subsection{Analysis of the Input Privacy}

In DLCD settings, the master reveals both global model and input data to the worker nodes. After these nodes locally train their models, they also reveal them to the master. However, in our proposal the master encodes the input 
data using PBACC, so it only shares the encoded version with the nodes. Since the nodes train their local 
models with the encoded data, the updated trained models are also in the encoded 
domain, so they are not directly transmitted to the master in clear. Hence, in 
this configuration, the global model constructed by the master is the only element
revealed to the nodes. Thus, only attacks 
over the global model would be possible, which are less harmful than attacks 
over the  input data or local models. As an example to illustrate this, a 
membership inference attack over a global model might reveal if some data element 
was used in the training set, but it does not identify which node owns it.

In DLDD scenarios, datasets are private, but the nodes reveal their local models to the master 
(i.e., aggregator) so it can aggregate a global model, and additionally, the 
master will reveal this global model to the nodes so they can train their local 
models. Consequently, the master has access to the local models and attacks, like membership inference, can now identify the owner
of the data element that was used to train the model. Most 
of the input privacy concerns are solved by our proposal of secure aggregation, 
where local models are encoded and exchanged to securely compute some
arbitrary aggregation function. This implies that the global model is the only 
element leaked to the nodes, so the privacy concerns are smaller than in non-secured FL. 
It should also be noted that, depending on the aggregation function, the
aggregated model reveals less information from the local models than having them in 
clear. This fact also affects attacks like membership inferences based on shadow 
models, as it will be harder for them to learn to identify if data elements
were used or not in the training phase. However, our secure training approach 
for DLDD is even better privacy-wise, as we ensure the input privacy 
of all the data elements treated in the configuration. Since the master encodes 
the global model using PBACC, this object is not reveal to the nodes either. 
Similarly, as the nodes train their local models taking an encoded global model 
as a basis, the resulting computed models are also in the encoded domain, so they are 
not known to the master. Additionally, the master can only decode an aggregated model 
which provides some degree of output privacy (see Section~\ref{sec:results}), 
making clear that this configuration is the most secure option. 
Table~\ref{table:input_privacy_comparison} lists the features of the presented comparison.

\begin{table*}[h!]
    \centering
    \small
    \caption{\label{table:input_privacy_comparison} Comparison of the sensible elements and the entity to which is revealed in both distributed and DLCD scenarios.}
    \begin{tabular}{lccc}
    \textsc{Scenario} & \textsc{Input} & \textsc{Local Model} & \textsc{Global Model}  \\ \hline
    Uncoded DLCD & Public (nodes) & Public (master) & Public (nodes) \\
    Secure training & Private & Private & Public (nodes) \\ \hline
    Uncoded DLDD & Private & Public (master) & Public (nodes) \\
    Secure aggregation & Private & Private & Public (nodes) \\
    Secure training & Private & Private & Private \\ \hline
    \end{tabular}
\end{table*}

\subsection{Efficiency Analysis}

For the centralized scenarios, uncoded decentralized learning has a communication 
cost of $N$ messages of size $\frac{L}{N}$ once, corresponding to the sharing of 
the split dataset to each of the nodes, plus $2N$ messages of the model size $W$ 
each round, corresponding to the exchange of the global model from the master to 
the nodes  ($N$ messages), and the exchange of the local models from the nodes to 
the master ($N$ additional messages). Regarding computation, the whole task is 
composed of training the local machine learning models by the nodes, and performing 
the aggregation operation to construct the global model by the master. Our proposal 
for secure training increases the communication cost to $N\frac{L}{K}$ messages of 
size $W$ in each round, and $N\frac{L}{K}$ of the model inference size ($B$), as 
the nodes require from the master to decode the result of the model execution and 
apply the optimization step, as discussed in~\ref{sec:integrating_pbacc}. This implies 
that for each model execution of an encoded batch, there are $N$ messages exchanged 
from the master to the nodes with the global model, and $N$ messages from the nodes 
to the master with the outcomes. Additionally, $N$ messages of size $\frac{L}{K}$ 
are sent once to share the encoded inputs. Regarding computation load, only one 
encoding of the dataset has to be done, and can be reused in several experiments, 
while $\frac{L}{K}$ decoding operations of size $K$ have to be done so the master 
can optimize the model after each distributed model execution.

For decentralized scenarios, normal FL has the same communication and 
computation cost of the DL case, but without the input sharing messages. Our 
proposal for secure aggregation increases the communication cost in $N(N-1)$ messages 
of the model size $W$ divided by $K$ each round, required to perform the secure 
aggregation. This messages are the result of each node having to exchange an encoded 
share of its local trained model with every other node. However, as the aggregation 
is then computed over the received shares, the obtained result is then sent to the 
master so it can decode the global model, which reduces the size of the $N$ messages 
required to exchange the model from the the nodes to the master, to $\frac{L}{K}$
size. The computation cost, in this case, is increased by an additional encoding 
operation of the model size $W$ and a decoding operation of $\frac{W}{K}$, both 
performed by all of the nodes in each round. Although the efficiency of the secure
aggregation is already decent, especially in terms of computational cost, it is much 
more improved in the second approach (secure training). In this case, the 
communication cost remains the same as in normal FL, as the encoded model has the 
same size of the original model, and the computation cost is only increased by an 
encoding operation and a decoding operation of the model size $W$, both performed by
the master. Additionally, it also removes the computational cost of the aggregation,
since this operation is done directly within the decoding step. The conclusions of 
this analysis are contained for reference in Table~\ref{table:communication_comparison}. 

\begin{table*}[h!]
    \centering
    \footnotesize
    \caption{\label{table:communication_comparison} Communication comparison of both distributed and DLCD scenarios.}
    \begin{tabular}{lp{0.3\textwidth}p{0.42\textwidth}}
    \textsc{Scenario} & \textsc{Communication cost} & \textsc{Computation cost} \\ \hline
    Uncoded DLCD &  $2N \times W$ per round & Training and aggregation each round \\
    Secure train. & $N\frac{L}{K} \times (W + B)$ per round plus
    $N \times \frac{L}{K}$ once & One dataset encoding size $L$ + training, aggregation  
    and $\frac{L}{K}$ decodings size $K$ each round \\ \hline
    Uncoded DLDD & $2N \times W$ per round  & Training and aggregation each round \\
    Secure Agg. & $(2N + N(N-1)) \times W / K$ per round 
    &  Training, aggregation, encoding and decoding model size ($W$) each round \\
    Secure Train. & $2N \times W$ per round 
    &  Training, encoding and decoding of model size ($W$)
    each round \\ \hline
    \end{tabular}
\end{table*}

\section{Results}
\label{sec:results}

We have performed extensive experimental tests to demonstrate the viability of 
our proposals for secure DL and FL. Our three different configurations have been 
tested with two representative ML models: (i) a convolutional neural network (CNN) 
using  the MNIST dataset, a (ii) a Variational AutoEncoder using the Fashion MNIST 
dataset, and (iii) a Cox Regression using the METABRIC~\footnote{\href{https://ega-archive.org/studies/EGAS00000000083}{https://ega-archive.org/studies/EGAS00000000083}}
dataset, a public dataset for the gene expression in primary breast cancers. The 
secure approaches have been compared with their non-secure counterparts, 
so that it is easier to see the real cost of adding the different forms of privacy in 
terms of efficiency and model accuracy.

Since the measured privacy gives the estimated leakage per data element, we can 
reuse them for both networks. This also allows us to compare how the different 
secure approaches behave depending on the ML model used, which is very helpful to
understand how different architectures affect PBACC performance, and to assess if 
the scheme ready for real-world DL and FL. All configurations have been tested with 
an NVIDIA RTX 3090 of $24$ GB of RAM, for different network sizes. Since all
nodes are running on the same machine, in parallel, when the master node makes 
operations, it will have much more computational power at its disposal than the 
nodes. This matches with many common scenarios, where the server
nodes have much more computational power than their client nodes. 
Communication among entities is performed using remote procedure calls (RPC) 
over HTTP.

For every set of experiments, we have measured the convergence rounds required to 
achieve the value of that metric, the computation times of the different phases 
of PBACC and the training of the model, and the sharing times of whole distributed
task. These latter times comprise the serialization and parsing times of the
information that has to be exchanged, and the actual communication time. In all
the DLDD test cases, the dataset has been split equally among each 
participating node.

\subsection{Experiments with CNNs}

\begin{table*}[h]
    \centering
    \small
    \caption{\label{table:cnn-parameters} CNN Experiment Parameters}
    \begin{tabular}{lcccccccc} 
    \textsc{Scenario} & $N$ & \textsc{BS} & \textsc{E} & $\gamma$ & $K$ & $T$ & $\sigma_n$ & \textsc{Leakage} \\ \hline
    Uncoded DLCD & $50$ & $10$ & $1$ & $10^{-3}$ & n/a & n/a & n/a & n/a \\
    Secure Train. & $50$ & $10$ & $1$ & $10^{-3}$ & $10$ & $30$ & $30$ & $\leq 0.70$ bit \\ \hline
    Uncoded DLDD & $50$ & $10$ & $1$ & $10^{-3}$ & n/a & n/a & n/a & n/a \\
    Secure Agg. & $50$ & $10$ & $1$ & $10^{-3}$ & $1$ & $30$ & $10$ & $\leq 0.60$ bit \\
    Secure Train. & $50$ & $10$ & $1$ & $10^{-3}$ & $1$ & $30$ & $10$ & $\leq 0.60$ bit \\ \hline
    \end{tabular}
\end{table*}

The parameters chosen for the experiments with CNNs are listed in 
Table~\ref{table:cnn-parameters}.  Every configuration has a fixed number of 
nodes $N$, equal batch size BS and epochs $E$ so we can make a fair
comparison between the computational cost. The learning rate $\gamma$ 
has been chosen so it minimizes the convergence round, given 
the other parameters. In all the test cases, it was found out that 
$\gamma = 10^{-3}$ was the best option except in the secure training over DLCD, 
where a higher learning rate ($2.5 \times 10^{-2}$) improved the convergence 
speed considerably.  This difference in behaviour is clearly related with how 
the aggregation is done in this configuration, which is not implicitly calculated, 
but obtained from the Berrut rational interpolation of each encoded local model. 
One possible interpretation is related to the fact that having lower learning 
rates increases the likelihood to avoid local minima, which makes the convergence 
slower. Another well-known way of avoiding local minima and helping in 
generalization involves adding some controlled noise, as in differential 
privacy schemes. As the native aggregation is introducing a bounded noise in 
the decoded model, increasing the learning rate also increases the convergence 
speed without inducing the loss function  to find local minima.

The number of random coefficients $T$ and the standard deviation $\sigma_n$ have
been chosen to ensure that the leakage is less than, at least, $1$ bit for any
group of $10$ honest-but-curious nodes ($20$\% of total). Each secure configuration 
has $T = 30$ random coefficients, and $\sigma_n = 10$, except the secure training 
over DLCD, which requires higher security ($\sigma_n = 30$) due to $K$ 
being higher.

\begin{table*}[h]
    \centering
    \footnotesize
    \caption{\label{table:cnn} Performance and computing times for CNN per round. $T_c$ is the average convergence round for each scheme, $A$ is the accuracy}
    \begin{tabular}{lcc|ccccc}
        & \multicolumn{2}{c}{\textsc{Convergence}} & \multicolumn{5}{c}{\textsc{Running times}} \\ \cline{2-3} \cline{4-8}
        \textsc{Scenario} & $T_c$  & $A$ & \textsc{Enc.} & \textsc{Shar.} & \textsc{Comp.} & \textsc{Dec.} & \textsc{Total} \\ \hline
        Uncoded DLDD & $13.6$ & $0.98$ & n/a & $3.31$ s & $3.32$ s & n/a & $80.88$ s \\
        Secure Agg. & $12.2$ & $0.98$ & $12.37$ s & $38.76$ s & $3.48$ s & $87.5 \mu$s & $742.70$ s \\
        Secure Train. & $47.1$ & $0.86$ & $0.02$ s & $29.29$ s & $7.30$ s & $30.06 \mu$s & $1724.33$ s \\ \hline
        Uncoded DLCD & $11.8$ & $0.98$ & n/a & $3.33$ s & $3.32$ s & n/a & $83.91$ s \\
        Secure Train. & $1.0$ & $0.98$ & $111.89^*$ s & $6449.87$ s & $268.65$ s & $4.3$ ms &$6830.85$ s \\ \hline
        {\footnotesize ${}^*$ Only once.} & & \multicolumn{1}{c}{} & & & & &
    \end{tabular}
\end{table*}
\begin{table*}[h]
    \centering
    \footnotesize
    \caption{\label{table:cnn_convergencies_security} Convergences of CNN with $T=30$ and different values of $\sigma_n$}
    \begin{tabular}{lcccc}
    \textsc{$\sigma_n$} & Secure Agg. (DLDD) & Secure Train. (DLDD) & Secure Train. (DLCD)\\ \hline
    $10$ & $0.98 \, (13.6)$ & $0.86 \, (47.1)$ & $0.98 \, (1)$ \\
    $50$ & $0.95 \, (17.1)$ & $0.73 \, (100.3)$ & $0.98 \, (2)$ \\
    $100$ & $0.86 \, (30.2)$ & $0.45 \, (78.9)$ & $0.98 \, (3)$ \\
    $200$ & $0.83 \, (64.9)$ & $0.24 \, (45)$ & $0.98 \, (4)$ \\
    $400$ & $0.83 \, (162.5)$ & $0.10 \, (16.7)$ & $0.94 \, (4)$ \\ \hline
    {\footnotesize $(\cdot)$ Average convergence round} & & &
    \end{tabular}
\end{table*}
\begin{figure*}[h!]
    \centering
    \includegraphics[width=0.8\linewidth]{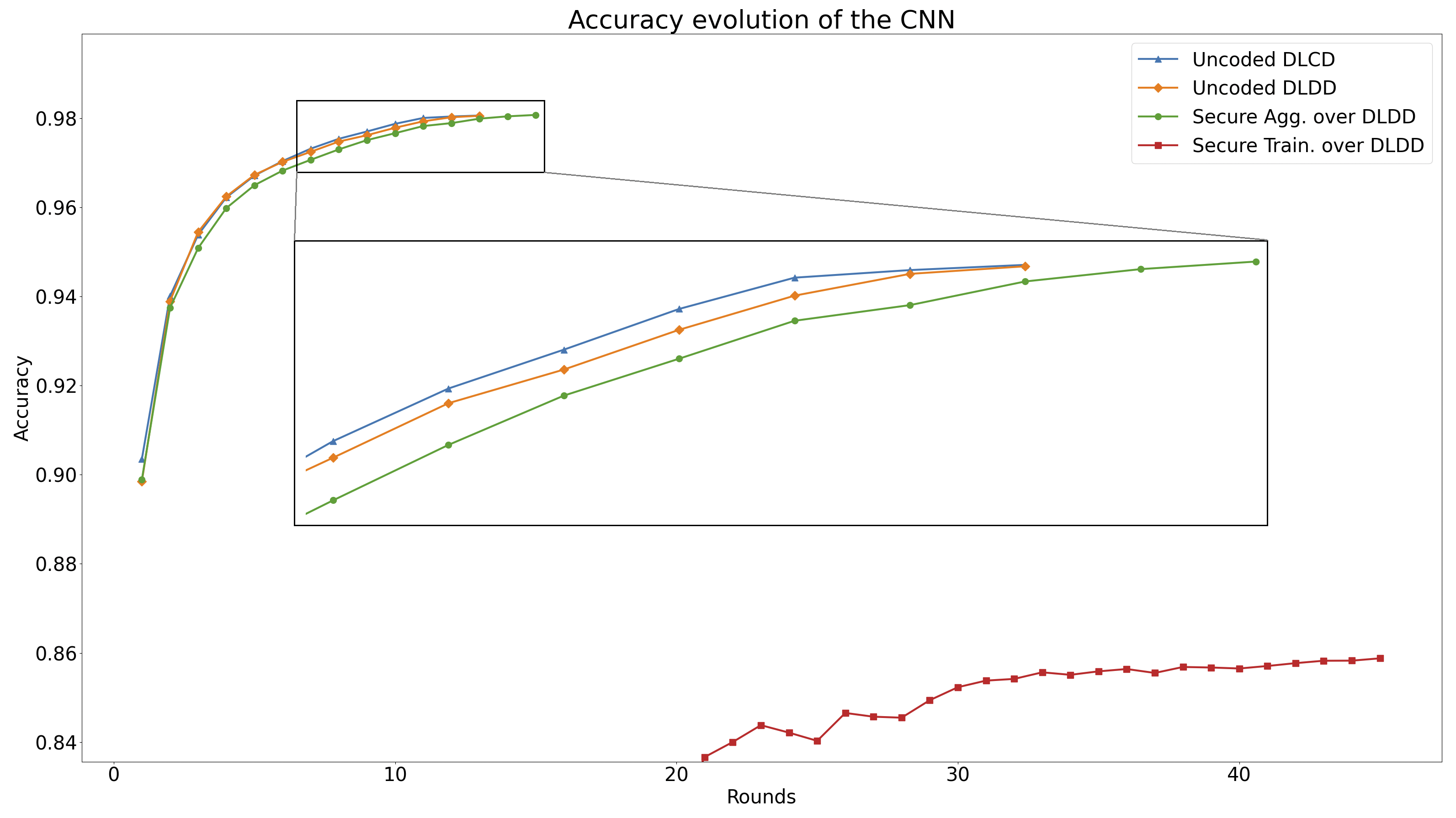}
    \caption{Comparison of the Accuracy evolution of all scenarios for the CNN experiments}
    \label{fig:cnn_secure_scenarios_comparison}
\end{figure*} 

The first interesting finding about the CNN tests reported in Table~\ref{table:cnn}
is that all configurations  achieve the same accuracy, except the secure training 
over DLDD, which converges at a lower value. Therefore, for almost all
configurations and this security threshold, the bounded error introduced by 
PBACC allows the CNN to converge without deteriorating the final accuracy. 
Additionally, the configuration that actually diminishes the final accuracy by 
a $12\%$ is the one with a higher level of privacy, since the end nodes 
do not see the global model. This accuracy loss is related to the noise introduced 
by this native aggregation performed at the decoding step, and it implies 
that the decoded global model provides some degree of output privacy. However, 
it would be necessary to implement membership inference attacks over the different 
trained models to confirm this fact~\cite{hu2022membership}.

Given that each configuration distributes the computation of 
different functions, it can be concluded that the errors of approximating single
functions are correlated instead of accumulated. Another interesting fact is the 
impact of the error introduced by PBACC in the convergence round. We observe, in 
the secure aggregation DLDD case, that the errors introduced by PBACC have almost 
no impact on the convergence round compared to the uncoded DLDD and DLCD. In 
contrast, the secure training over DLCD only requires one round to converge,
since this configuration is equivalent, from a model
training perspective, to a normal centralized training by a single entity. 
As with accuracy, the outlier is the secure training over DLDD, where the 
convergence increases by approximately $35$ rounds. Although the
convergence is slower, it still reaches a good accuracy thanks to the 
adjustment of the learning rate.

The measured results for the training time match the theoretical expected 
behaviour for all configurations. Uncoded DLDD and DLCD present similar values 
in the sharing and training phases. The first one corresponds to 
the exchange of the global model and the local models, and the second one corresponds 
to the computation of the training task. In both cases, the aggregation of the model 
has a negligible value in comparison with the rest of the computation times. 
Uncoded DLCD has an additional $5.44 s$ of input sharing time that only happens 
once, corresponding to the exchange of the split dataset to the worker nodes.
Looking at the DLDD scenarios, we see that the secure aggregation does not affect the
computation task, as the training has a similar cost as in the uncoded DLDD, and the
aggregation done by the nodes is not significant. However, it introduces a new 
encoding operation in the nodes that adds an overhead ($12.37$ s), and a very 
fast decoding step done by the master. Regarding communication, the main overhead
introduced appears on the sharing phase, as it now includes the sharing of the
encoded models among nodes. Comparing it with the secure training approach, 
it is more efficient from the perspective of a DLDD round, as the only overheads
introduced are an encoding and a decoding step done by the master. Although 
the communication time of this configuration is similar to the uncoded
DLDD, in this case the aggregation has to serialize $N$ models before sharing, 
instead $1$.

Looking at the secure training over DLCD, we see that the overheads are noticeably
higher. The encoding of the input dataset is much slower than encoding the model 
parameters, but it only has to be done once, and the dataset can be reused for
different experiments. The encoded dataset has to be also shared once with the 
nodes ($16.85$ seconds) in contrast with the input sharing done in uncoded 
DLCD ($5.45$ seconds). As the secure training is iterative, we see that the 
repetitive sharing and decoding phases have increased these times ($6449.87$ s and
$4.3$ ms respectively). It has also increased the training time, since in 
uncoded DLCD each node has $\frac{L}{N}=1200$ entries, whereas in secure training
this raises to $\frac{L}{K}=6000$ entries.
Combining the measurements of both times and convergence rounds the overall 
behaviour shows that, for DLDD scenarios, secure training is more efficient 
per round, but the time is greater than with secure aggregation. 
Figure~\ref{fig:cnn_secure_scenarios_comparison} depicts the evolution
of the model accuracy over the training rounds.

We also analyzed the convergence of the model as a function of the target 
privacy leakage. We set $T=30$ random coefficients and ran experiments for 
different values of the standard deviation of the noise 
$\sigma_n = \{10, 50, 100, 200, 400\}$. The results can be observed in 
Table~\ref{table:cnn_convergencies_security}. As we increase the level of
privacy in secure DLDD, the precision decreases and convergence is slower, yet
the utility of the model is still reasonable with very high privacy levels. 
These precisions could be improved if we increase $K$, as in the DLCD case. 
Thus, most of the cost of adding privacy is related to an overhead of the 
convergence round. 

For secure training over DLDD, the quality of the model reduces 
substantially as $\sigma_n$ increases. This is consistent with the overall 
behaviour of this configuration, and suggest that training is much more sensitive 
to noise perturbations. It is also clear that having $K=1$ affects the precision
of this configuration, but contrary to what happens in the secure aggregation,
we cannot increase this value to improve it due to the training function being
approximated.
In secure training over DLCD, PBACC is very robust 
in precision. This behaviour might seem surprising, as this case deals with a 
more complex function than the secure aggregation over DLDD. The explanation
is twofold: (i) $K=10$ in this case, which clearly favors the decoding 
operation thanks to the well distributed points, and (ii) this is the only 
configuration where the results obtained from the master are composed from 
evaluations of the same function.

\subsection{Experiments with Variational Autoencoders}

To demonstrate the validity of our encoded scheme with other ML models, we
tested the system with a categorical VAE based on the Gumbal-Softmax 
reparametrization. The reason for avoiding the vanilla  VAE is that its 
reparametrization becomes very unstable for the Secure training scenario over DLDD. 
In that configuration, the model parameters are encoded using PBACC, which leads to 
the internal encoder of the VAE to generate values in the encoded domain. As a result.
the vanilla VAE must compute some exponential functions of large value, causing 
numerical overflow. The actual parameters chosen for the VAE experiments are listed 
in Table~\ref{table:vae-parameters}. We considered in this case a smaller network 
of $N = 30$ nodes, so this will  allow us to evaluate the behaviour of the scheme
with less worker nodes but with higher computational power. The learning rate $\gamma$ 
was picked to minimize the convergence round. As with CNNs, we empirically found 
that $10^{-3}$ was the best choice, except for the secure training over DLDD.


\begin{table*}[h]
    \centering
    \small
    \caption{\label{table:vae-parameters} VAE Experiment Parameters}
    \begin{tabular}{lcccccccc}
    \textsc{Scenario} & $N$ & \textsc{BS} & \textsc{E} & $\gamma$ & $K$ & $T$ & $\sigma_n$ & \textsc{Leakage} $\epsilon$ \\ \hline
    Uncoded DLDD & $30$ & $10$ & $1$ & $10^{-3}$ & n/a & n/a & n/a & n/a \\
    Secure Agg. & $30$ & $10$ & $1$ & $10^{-3}$ & $1$ & $18$ & $10$ & $\leq 1.0$ bit \\
    Secure train. & $30$ & $10$ & $1$ & $2.5 \cdot 10^{-2}$ & $1$ & $18$ & $10$ & $\leq 1.0$ bit \\ \hline
    Uncoded DLCD & $30$ & $10$ & $1$ & $10^{-3}$ & n/a & n/a & n/a & n/a \\
    Secure train. & $30$ & $10$ & $1$ & $10^{-3}$ & $10$ & $18$ & $30$ & $\leq 1.0$ bit \\ \hline
    \end{tabular}
\end{table*}
\begin{figure*}[h]
    \centering
    \includegraphics[width=0.8\linewidth]{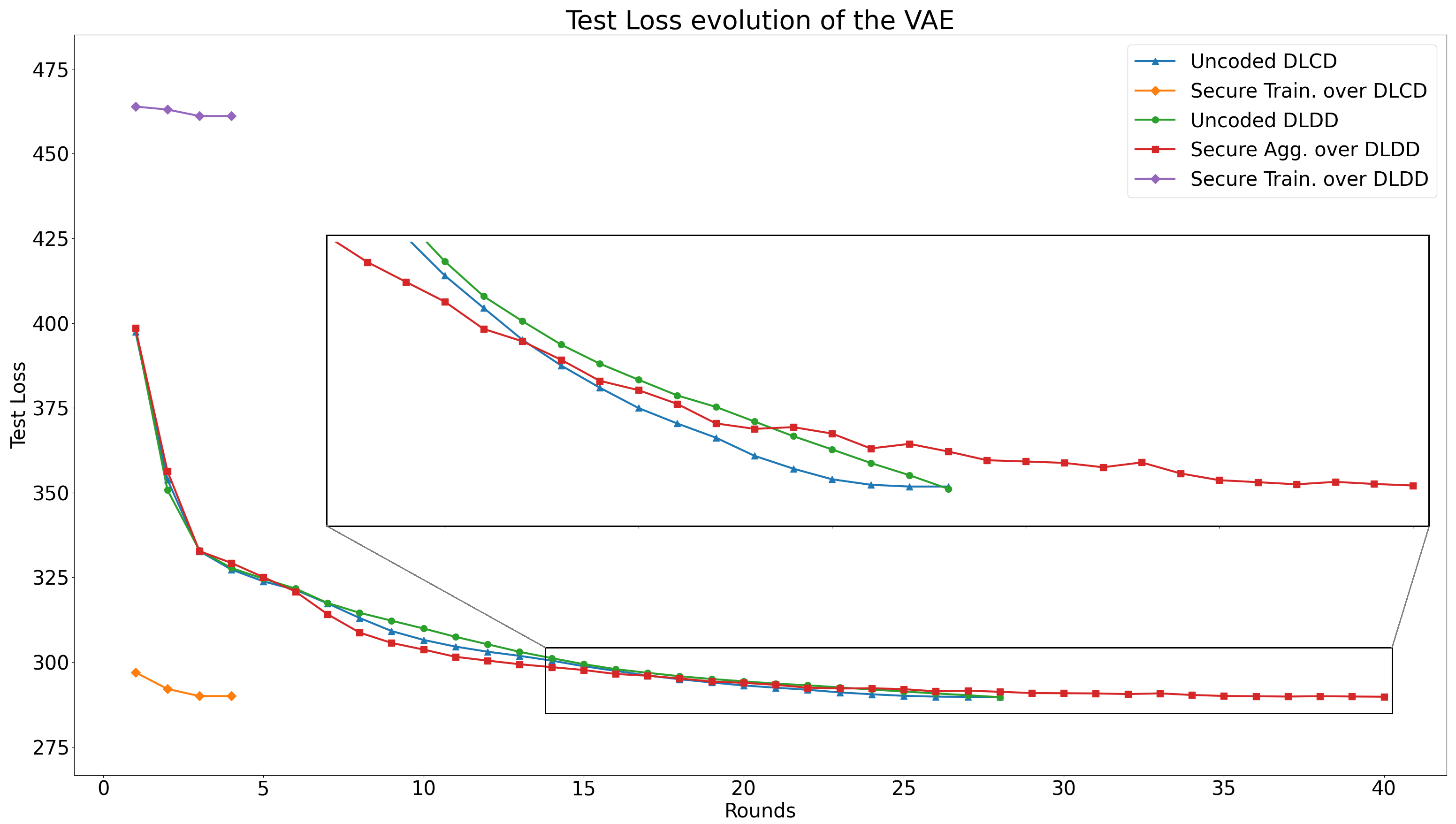}
    \caption{Comparison of the Accuracy evolution of all scenarios for the VAE experiments}
    \label{fig:vae_accuracies_evolution}
\end{figure*} 
\begin{table*}[h]
    \centering
    \footnotesize
    \caption{\label{table:vae} Performance and computing times for VAE per round. $T_c$ is the average convergence round for each scheme, $L$ is the test loss}
    \begin{tabular}{lcc|ccccc}
    & \multicolumn{2}{c}{\textsc{Convergence}} & \multicolumn{5}{c}{\textsc{Running times}} \\ \cline{2-3} \cline{4-8}
    \textsc{Scenario} & $T_c$  & $L$ & \textsc{Enc.} & \textsc{Shar.} & \textsc{Comp.} & \textsc{Dec.} & \textsc{Total} \\ \hline
    Uncoded DLDD & $25.8$ & $289.77$ & n/a & $40.40$ s & $19.46$ s & n/a & $1544.39$ s \\
    Secure Agg. & $35.8$ & $289.81$ & $20.12$ s & $140.50$ s & $20.26$ s & $0.601 \mu$s & $6475.50$ s \\
    Secure Train. & $2.6$ & $461.07$ & $0.027$ s & $104.21$ s & $21.45$ s & $0.82 \mu$s & $326.79$ s \\ \hline
    Uncoded DLCD & $26.2$ & $289.66$ & n/a & $36.11$ s & $18.47$ s & n/a & $1430$ s \\
    Secure Train. & $2$ & $290.00$ & $50.12^* s$ & $62904.54$ s & $1907.81$ s & $0.18$ s & $129675.18$ s \\ \hline
    {\footnotesize ${}^*$ Only once.} & & \multicolumn{1}{c}{} & & & & &
    \end{tabular}
\end{table*}

Like in the CNN test, we see (Table~\ref{table:vae}) that 
the VAE converges to a similar test loss, except in the case of secure training 
over DLDD, where it is much greater. This confirms the idea that this configuration
is providing some degree of output privacy since, from a theoretical perspective, 
a variational auto encoder is a model that learns the distribution of the data which 
is being trained with, a task similar to what an attacker seeks in a membership
inference attack with a shadow model~\cite{hu2022membership}. Recall 
the high complexity of the function being computed in this configuration, 
where the scheme has to approximate several rounds of the encoding, 
reparametrization, decoding, loss function calculation and optimization step.
The convergence round show a pattern also similar to that with CNN, suggesting 
that errors introduced by PBACC do not have a  significant impact on the evolution
of the secure configurations, with the exception of secure training over DLDD. 
This configuration seems to stop learning too early, and the resulting quality is 
low. The secure training over DLCD again requires only $2$ to learn the model,
thanks to the training task being done as if the master was the only one
executing this task.

\begin{table*}[h]
    \centering
    \small
    \caption{\label{table:cvae_convergencies_security} Convergences of VAE with $T=30$ and different values of $\sigma_n$}
    \begin{tabular}{lcccc}
    \textsc{$\sigma_n$} & Secure Agg. (DLDD) & Secure Train. (DLDD) & Secure Train. (DLCD)\\ \hline
    $10$ & $289.81 \, (35.8)$ & $461.07 \, (2.6)$ & $289.77 \, (2)$ \\
    $50$ & $371.95 \, (7.2)$ & $608.86 \, (3.3)$ & $290.00 \, (3)$ \\
    $100$ & $736.07 \, (7.1)$ & $614.29 \, (4.7)$ & $289.18 \, (4)$ \\
    $200$ & $7084.40 \, (1)$ & $946.40 \, (9.8)$ & $290.08 \, (8)$ \\
    $400$ & $8980.47 \, (15.4)$ & $23361.69 \, (14.5)$ & $293.60 \, (25.7)$ \\ \hline
    {\footnotesize $()$ Convergence round} & & &
    \end{tabular}
\end{table*}

Regarding the time measurements of the  different computation and communication phases 
for this experiment, uncoded DLDD and DLCD present similar values in the sharing
and training phases, as expected. The values are higher than for the CNNs, though
this is compensated by shorter training times. The aggregation times continue to be
negligible for the non-secure configurations, while the uncoded DLDD adds an 
additional overhead due to the exchange of the initial dataset among workers.

Looking at the DLDD scenarios, we see that the secure aggregation 
presents a similar behavior to that of CNN, where the encoding presents an 
overhead of $20.12$ s due to being held in the workers, the sharing time is the 
greatest overhead with a value of $140.50$ s, the training is not being affected, 
and the decoding adds a small value of $0.601 \mu$s. In contrast, the secure 
training approach only provides a significant overhead in sharing phase ($104.21$ s), 
due to having to serialize $N$ models, and the encoding and decoding steps increase 
small times ($0.027$ s and $8.20 \mu$s). Here, we can start to observe
the previous statement about this configuration behaving better, with more powerful 
nodes, than the secure aggregation. It is easy to check that the secure training is 
$34.82\%$ faster than the secure aggregation in the sharing phase, while it was
a $32.33\%$ in the CNN experiment. Looking at the secure training over DLCD, 
the encoding of the input dataset is faster by the smaller network size. For the VAE, 
the repetitive sharing and decoding phases have a greater increase mainly due to the
greater size of the model. However, the training task is proportionally reduced.
Note that in this case each node has $\frac{L}{N} = 2000$ entries, and, in secure
training, each node has $\frac{L}{K} = 6000$ entries. Figure~\ref{fig:vae_accuracies_evolution} depicts the evolution of the test loss over
the rounds of the training task.


We also analyzed the convergence of the VAE in function of the target security level.
To that end, we set $T=18$ random coefficients and ran the experiments for different
values of noise, $\sigma_n = \{10, 50, 100, 200, 400\}$. The results are collected in
Table~\ref{table:cvae_convergencies_security}. Secure aggregation over DLDD, turns
out to be highly sensitive to the error as the level of privacy increases. the model is much more sensitive to the error introduced than in the CNN case but, unlike the CNNs,
we do not see degradation on the utility. For the secure aggregation over DLDD, 
convergence is reached earlier. The secure training over DLDD is consistent with the
results obtained previously. In secure training over DLCD, PBACC is very robust 
in precision, since almost all levels of security tested converge with a test loss 
of $290$. The only one that presents slightly worse results is the one tested with 
$\sigma_n=400$, but the test loss obtained is still quite similar. Here, the cost of
adding this privacy is visible in the convergence round.

\subsection{Survival Analysis}

\begin{table*}[h!]
    \centering
    \small
    \caption{\label{table:cox-parameters} COX Experiment Parameters}
    \begin{tabular}{lcccccccc} 
    \textsc{Scenario} & $N$ & \textsc{BS} & \textsc{E} & $\gamma$ & $K$ & $T$ & $\sigma_n$ & \textsc{Leakage} $\epsilon$ \\ \hline
    Normal FL & $70$ & $10$ & $10$ & $10^{-2}$ & n/a & n/a & n/a & n/a \\
    Secure Aggregation & $70$ & $10$ & $10$ & $10^{-2}$ & $1$ & $42$ & $10$ & $\leq 0.60$ bit \\
    Secure training (dec.) & $70$ & $10$ & $10$ & $10^{-2}$ & $1$ & $42$ & $10$ & $\leq 0.60$ bit \\ \hline
    Normal DL & $70$ & $10$ & $10$ & $10^{-2}$ & n/a & n/a & n/a & n/a \\
    Secure training (cent.) & $70$ & $10$ & $10$ & $10^{-2}$ & $10$ & $42$ & $30$ & $\leq 0.70$ bit \\ \hline
    \end{tabular}
\end{table*}
\begin{figure*}[h!]
    \centering
    \includegraphics[width=0.8\linewidth]{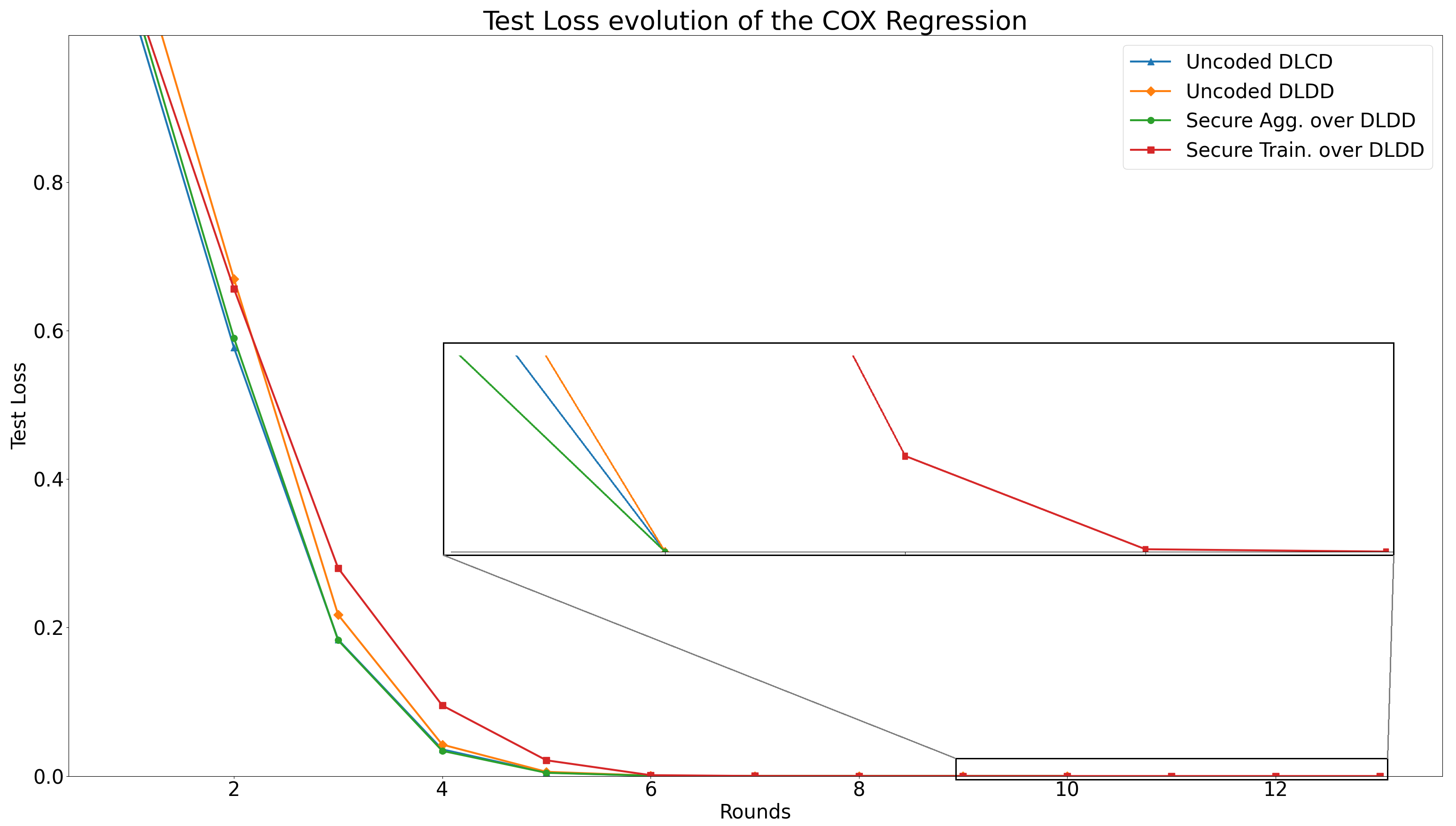}
    \caption{Comparison of the Accuracy evolution of all scenarios for the COX experiments}
    \label{fig:cox_accuracies_evolution}
\end{figure*}
\begin{table*}[h!]
    \centering
    \footnotesize
    \caption{\label{table:cox} Performance and computing times for COX Regression per round. $T_c$ is the average convergence round for each scheme, $L$ is the test loss}
    \begin{tabular}{lcc|ccccc}
    & \multicolumn{2}{c}{\textsc{Convergence}} & \multicolumn{5}{c}{\textsc{Running times}} \\ \cline{2-3} \cline{4-8}
    \textsc{Scenario} & $T_c$  & $L$ & \textsc{Enc.} & \textsc{Shar.} & \textsc{Comp.} & \textsc{Dec.} & \textsc{Total} \\ \hline
    Uncoded DLDD & $10$ & $0.0$ & n/a & $2.61$ s & $3.59$ s & n/a & $62$ s \\
    Secure Agg. & $10$ & $0.0$ & $17.34$ s & $38.68$ s & $3.82$ s & $1.10 \times 10^{-4}$s & $598.40$ s \\
    Secure Train. & $13.6$ & $0.0$ & $0.02$ s & $4.26$ s & $3.43$ s & $4.9 \times 10^{-4}$ s & $104.86$ s \\ \hline
    Uncoded DLCD & $10$ & $0.0$ & n/a & $2.49$ s & $3.67$ s & n/a & $61.6$ s \\
    Secure Train. & $1$ & $0.0$ & $1.88^*$ s  & $15.23$ s & $21.10$ s & $3.6 \times 10^{-4}$ s & $38.21$ s \\ \hline
    {\footnotesize ${}^*$ Only once.} & & \multicolumn{1}{c}{} & & & & &
    \end{tabular}
\end{table*}
\begin{table*}[h!]
    \centering
    \small
    \caption{\label{table:cox_convergencies_security} Convergences of VAE with $T=30$ and different values of $\sigma_n$}
    \begin{tabular}{lcccc}
    \textsc{$\sigma_n$} & Secure Agg. (DLDD) & Secure Train. (DLDD) & Secure Train. (DLCD)\\ \hline
    $10$ & $0.0 \, (10)$ & $0.0 \, (13.6)$ & $0.0 \, (1)$ \\
    $50$ & $0.0 \, (10)$ & $0.73 \, (1)$ & $0.0 \, (1)$ \\
    $100$ & $9.38\cdot 10^{-10} \, (15.8)$ & $4.45 \, (1)$ & $0.0 \, (1)$ \\
    $200$ & $3.75\cdot 10^{-9} \, (12.6)$ & $4.60 \, (1)$ & $0.0 \, (1)$ \\
    $400$ & $2.01 \cdot 10^{-7} \, (10.5)$ & $5.49 \, (1)$ & $0.0 \, (1)$ \\ \hline
    {\footnotesize $()$ Convergence round} & & &
    \end{tabular}
\end{table*}

The last model chosen is a Cox-Time regression model from~\cite{coxtime2019}. 
The parameters used for this COX regression experiments can be found in
Table~\ref{table:cox-parameters}. The model size is smaller than 
the others, and we test each configuration for a higher number of nodes.

Contrary to what happened in previous experiments, there is no difference in the 
final achieved quality of the model. This model is much easier to approximate than 
the previous ones, and even the secure training over decentralized shows a similar
convergence round compared to the other options. So, regardless the configuration, 
for this security threshold,  the bounded error introduced by PBACC allows the COX 
regression to converge without having an important impact on the final precision or 
the convergence round. Table~\ref{table:cox} shows the measured times. As this
is the smallest of the tested models, it has substantially less overheads in 
comparison. The results obtained are nevertheless  consistent with the  previous
experiments, with uncoded DLDD and uncoded DLCD showing similar performance in 
every phase of the round and negligible aggregation times; their secure counterparts
are only a little slower in some of the phases. 
Focusing now on the DLCD scenarios,  the secure training configuration
benefits from larger networks with smaller models. This means that this scenario will
behave better in cross-device FL rather than in cross-silo FL, as their overheads will be
less significant.

Looking at the overall behavior in Table~\ref{table:cox}, we see a 
similar tendency than in the VAE experiments, with the exception that
the secure training over DLCD even outperforms the normal configurations.
This occurs thanks to the sharing and computation phases being much more efficient 
in this configuration, allowing the training operation to compensate the slower 
learning given by the fact that the other configurations have to aggregate a
global model. Figure~\ref{fig:cox_accuracies_evolution} depicts the evolution of the 
model test loss over the rounds of the training task.

Analyzing the precision of the model for the different levels of privacy, 
we measured the best results from all the models. In
Table~\ref{table:cox_convergencies_security},
we see that all levels of privacy tested for the secure aggregation over DLDD 
provide test losses very close to $0$. Additionally, the convergence round increases
less than in the other models when increasing privacy. Secure training over DLDD 
presents a similar test loss and convergence round for $\sigma_n=10$ than in the 
other cases, but gets worse as we increase privacy. Secure training over DLCD
attains the best results in terms of the privacy precision exchange. 
The model converges at the same test loss and convergence round independently of the
noise added.

\section{Conclusions}
\label{sec:conclusions}

In this paper, we have extended the scope for the application of approximate coded computing (and particularly Private Berrut Approximate Coded Computing) to decentralized computing systems. The new encoding and decoding algorithms can be implemented either in centralized or in fully decentralized forms and arranged to work with tensors, and still
meet tight bounds on the privacy leakage metric. We presented in detail the application to secure aggregation and secure training, in distributed or centralized forms. Through
numerical experiments, we have demonstrated that the scheme is flexible and robust,
so that it can be incorporated into disparate learning models at scale. One advantage of
our method is that, opposite to other forms of added noise like differential privacy, the error does not accumulate with an increasing number of nodes. In contrast, one limitation of PBACC is that the model quality deteriorates if the privacy leakage threshold
is very close to $0$. Future work will address this issue along with the combination
of PBACC with algorithms for verify the computations.





\section*{Acknowledgments}
This work was partially funded by the European Union’s Horizon Europe
Framework Programme for Research and Innovation Action under project
TRUMPET (proj. no. 101070038), the grant PID2020-113795RB-C33
funded by MICIU/AEI/10.13039/501100011033 (COMPROMISE project),
the grant PID2023-148716OB-C31 funded by MCIU/AEI/ 10.13039/501100011033
(DISCOVERY project), "Contramedidas Inteligentes de Ciberseguridad para la
Red del Futuro (CICERO)" (CER-20231019) funded by CERVERA grants 2023 for RTOs,
partially funded by the European Union’s Horizon Europe Framework Programme for
Innovation Action under project TrustED (proj. no. 10.3030/101168467) and ”TRUFFLES: TRUsted
Framework for Federated LEarning Systems", within the strategic cybersecurity
projects (INCIBE, Spain), funded by the Recovery, Transformation and Resilience
Plan (European Union, Next Generation). Additionally, it also has been funded
by the Galician Regional Government under project ED431B 2024/41 (GPC).

Funded by the European Union. Views and opinions expressed are however those of
the authors only and do not necessarily reflect those of the European Union.
Neither the European Union nor the granting authority can be held responsible
for them.

\bibliographystyle{IEEEtran}
\bibliography{references}  






\end{document}